\documentclass[times, review, 10pt]{elsarticle}

\usepackage{eccvabbrv}

\usepackage{graphicx}
\usepackage{booktabs}

\usepackage[accsupp]{axessibility}  

\usepackage{caption}

\usepackage{color}
\usepackage{epsfig}
\usepackage{amsmath}
\usepackage{amssymb}
\usepackage{makecell}
\usepackage{subcaption}
\usepackage{arydshln}
\usepackage{multirow}
\usepackage{soul}
\usepackage{setspace}

\usepackage{appendix}
\usepackage{algorithm}
\usepackage{algpseudocode}

\title{\vspace{-2.2cm}DDS-NAS: Dynamic Data Selection within Neural Architecture Search \\via On-line Hard Example Mining applied to Image Classification}
\author {
	Matt Poyser\textsuperscript{\rm 1},
	Toby P. Breckon\textsuperscript{\rm 1,2} \\
	\textsuperscript{\rm 1}Department of Computer Science, Durham University, UK.\\
	\textsuperscript{\rm 2}Department of Engineering, Durham University, UK.
}

\begin{document}
	\maketitle
	\begin{abstract}
		In order to address the scalability challenge within Neural Architecture Search (NAS), we speed up NAS training via dynamic hard example mining within a curriculum learning framework. By utilizing an autoencoder that enforces an image similarity embedding in latent space, we construct an efficient \textit{kd}-tree structure to order images by furthest neighbour dissimilarity in a low-dimensional embedding. From a given query image from our subsample dataset, we can identify the most dissimilar image within the global dataset in logarithmic time. Via curriculum learning, we then dynamically re-formulate an unbiased subsample dataset for NAS optimisation, upon which the current NAS solution architecture performs poorly. We show that our DDS-NAS framework speeds up gradient-based NAS strategies by up to 27$\times$ without loss in performance. By maximising the contribution of each image sample during training, we reduce the duration of a NAS training cycle and the number of iterations required for convergence.
	\end{abstract}
	
	\section{Introduction} 
	\begin{figure}[!h]
		\small
		\centering
		\includegraphics[width=\linewidth]{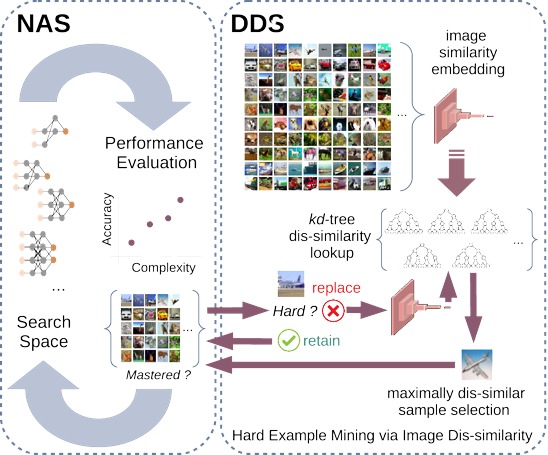}
		\caption{Overview of the DDS-NAS search phase. After a given training iteration, we determine whether a sufficient percentage of the data in the current subset is correctly classified, according to some \textit{a priori} \textit{mastery} threshold. If the subset has been mastered, we reformulate it dynamically. \textit{Hard} images in the current subset are retained, according to some \textit{a priori} \textit{hard}-ness threshold, while easy images are replaced with the most different image from the same class. To determine the most different image, we employ an (approximate) furthest-neighbour \textit{kd}-tree whereby each image is represented by the auto-encoded representation of its features within the latent space.}
		\label{fig:flow}
	\end{figure}
	
	Following the emergence of big data and the ever-increasing public availability of datasets, each with tens of thousands of data points, research within the deep learning domain is accelerating \cite{dltrends2}. Consequently, there are two key factors that need to be addressed. Firstly, the process by which we present data to the deep learning model is paramount. It is not uncommon for models to be trained for thousands of epochs, and thus any superfluous data within the dataset will have an increasingly negative impact on training speed. This phenomenon has given rise to hard example mining \cite{hemoriginal}, which attempts to identify \textit{hard} images (i.e. images that contribute highly to loss, upon which the model performs poorly). By only considering these hard images, we can not only sample from a minimal dataset, therefore minimising the duration of a training epoch, but also reduce the number of iterations required for model convergence, as the contribution of each image sample is maximised in every iteration.
	
	Similarly, the images sampled by the model in any given training iteration are controlled via curriculum learning \cite{curriculumOrig} and self-paced learning \cite{selfPaced}. Contrary to hard example mining, in which commonly only a subset of the global dataset is considered during the entire training process, curriculum learning and self-paced learning forces the initial iterations to sample one fraction of the global dataset, and subsequent iterations to sample from different fractions, until the entire global dataset is considered. Generally, curriculum learning introduces harder images (pre-defined by prior knowledge) as training progresses, while self-paced learning determines the current model performance as feedback to the controller to determine which images to sample next. 
	
	The second challenge arising from data accessibility is the evolution of the architecture search space. As research within the domain continues, newer, and often more complex network architectures are presented. To overcome this notion Neural Architecture Search (NAS) has emerged, which automatically traverses the architecture search space for a given task and generates models that are competitive alongside hand-crafted, state-of-the-art models \cite{naslitpoyser}. We can divide the NAS domain into evolutionary, reinforcement-learning, prediction-based, and gradient-based NAS frameworks. This paper primarily considers gradient-based NAS frameworks.
	
	More precisely, the seminal gradient-based DARTS \cite{darts} framework constructs a super-network, in which each layer consists of all possible operations in the search space, followed by a softmax layer across said operations, such that operation selection can be represented as (continuous) operation-magnitude optimisation. After training the super-network, the best-performing subset of operations are extracted, thus formulating a cell (sub-network). A series of these cells is then trained to generate a final `searched' model, fine-tuned upon a given challenge dataset.
	
	We propose a strategy that incorporates a novel combined hard example mining and curriculum learning approach to enable Dynamic Data Selection (DDS) within a NAS framework, denoted as DDS-NAS. By using image similarity as a proxy metric for image difficulty (on an easy to hard performance axis), we can select \textit{hard} images for processing within a given NAS training iteration in logarithmic time without compromising image diversity (Fig. \ref{fig:flow}). This process allows us to significantly improve the NAS search phase speed. Whilst this paper specifically addresses image datasets, there is no reason not to apply identical techniques to other application domains such as natural language processing (NLP). 
	
	On this basis, our main contributions are as follows:
	\begin{itemize}
		\setlength\itemsep{-0.1cm}
		\item[--] a novel framework, DDS-NAS, that incorporates both hard example mining and curriculum learning in order to minimise the training duration of a given epoch within NAS, demonstrated to be effective across a variety of commonplace NAS approaches (DARTS \cite{darts}, P-DARTS \cite{pdarts} and TAS \cite{tas}).
		\item[--] an efficient and novel approach for hard example mining within the image domain, that considers image dissimilarity an alternative metric to hardness, and employs an autoencoder architecture that enforces an image similarity embedding in latent space. This yields efficient dissimilar image look-up from a \textit{kd}-tree structure.
		\item[--] generation of models in a manner intrinsically robust to biased datasets, and 10 times quicker than existing NAS techniques, whilst retaining competitive, near state-of-the-art accuracy with minimal memory footprint over common benchmarks. 
	\end{itemize}
	
	\section{Prior Work}
	\noindent
	In this section we introduce the related NAS, hard example mining, and curriculum learning approaches, from which we draw our methodology. We restrict our NAS literature survey to only a brief overview of NAS techniques, since DDS-NAS can be deployed upon any NAS approach that iteratively processes images (evolutionary, reinforcement-learning, and gradient-based). Through this review of the literature, we highlight our contribution to the field, outlining the ways in which our framework works alongside the current NAS approaches to optimise performance and reduce computational requirements. 
	
	\subsection{Neural Architecture Search}
	\noindent
	With the rise of NAS, a multitude of recent literature has addressed the scalability challenge which occurs due to the resultant large search space and training cost. Following the seminal work of Zoph et al. \cite{zoph} and other reinforcement learning \cite{reinforce1, reinforce3}, and evolutionary \cite{evo7, evo8} approaches to NAS, weight-sharing techniques \cite{enas} reduce the need to train each architecture in the search space separately. 
	
	\subsection{NAS Strategies}
	Gradient-based approaches \cite{darts, pdarts, dnal} enable the application of stochastic gradient descent and other well-used deep learning techniques by relaxing the search space so that it is continuous, thereby drastically improving the convergence rate of the architecture. One-shot NAS approaches employ the weight-sharing super-network training stage of DARTS, with an alternative sampling strategy, tending to consider only one path through the super-network in a given training iteration \cite{senas}.
	
	Progressive DARTS (P-DARTS) \cite{pdarts} address the optimisation gap within DARTS between the sub-network and final model. This is achieved by simultaneously limiting the prevalence of skip-connections within a generated cell and by progressively reducing the operation search space available to the super-network. This in turn enables progressively increasing network depth.
	
	Network Pruning via Transformable Architecture Search (TAS) \cite{tas} crafts a loss function to directly minimise the complexity of the searched network. To this end, both the width (number of channels in a layer) and the depth (number of layers) of the network are also searched. By employing the knowledge distillation algorithm from \cite{kd}, weights from the fully trained super-network can be transferred to the `pruned' searched network.
	
	\subsection{Curriculum and Coreset Sampling Within NAS}
	CNAS \cite{cnas} employs a curriculum learning framework within NAS architecture, in order to slowly introduce new operations to the NAS controller search space, allowing the model to successfully master harder tasks as training progresses. 
	Overall, network topology is the primary focus for contemporary NAS solutions \cite{tas,cnas, pdarts}. By contrast, only minimal consideration of the dataset presented within the NAS pipeline is present in the literature.
	
	CLOSE \cite{close} uses curriculum learning to modify the sharing extent. There is no effort to reduce the training dataset size, but image hardness and uncertainty (which can be calculated from a range of different sub-network outputs) is factored into the loss computation.  
	
	Peng et al. \cite{pinas} introduce negative samples within NAS training, drawing from the benefits of contrastive learning. Core-set Sampling \cite{coresetnas} select a small subset of the data space for training the NAS super-network via the greedy $k$-center algorithm. ADAPTIVE-NAS \cite{adaptivenas} compares different core-set sampling algorithms for PT-DARTS \cite{critic2}, including adaptive sampling, in which the training set is periodically updated using GLISTER \cite{glister}. While their work is most similar to ours, there is no effort to consider image hardness, and is thus unable to utilise any benefits of curriculum learning. Moreover, only one search algorithm is evaluated with core-set selection. However, the core-set selection algorithm depends upon embeddings that are well aligned with the training data, much like DDS-NAS (Table \ref{tab:ablauto}). 
	
	To our knowledge, this paper represents the first approach to jointly employ online hard example mining and curriculum learning during NAS learning to optimise both model performance and reduce overall NAS computation requirements. 
	With the variety and quickly evolving nature of NAS strategies, it is imperative that our method can be deployed alongside any existing NAS approach. Our work is thus the first to utilise a core-set approach in conjunction with a variety of existing NAS approaches and different architecture search spaces. Our approach is able to accelerate training for even the oldest NAS methods, for which training speed is a known drawback \cite{senas}.
	
	\subsection{Curriculum Learning and Hard Example Mining}
	\noindent
	Graves et al. \cite{curriculumSurrogate} posit the need for a surrogate measure of learning progress to inform the curriculum controller, rather than model accuracy. They introduce several different measures, identifying the best as prediction gain (instantaneous loss for a sample) and gradient variational complexity (using the direction of gradient descent to measure model complexity).
	
	Hachoen and Weinshall \cite{selfPaced} suggest instead to use a scoring function to generate the curriculum. The scoring function ranks images within the dataset by difficulty through testing either the same model (pre-trained without curriculum learning) or a different model. Harder images are introduced to the model over time. Weinshall et al. \cite{curriculumTaskDifficulty} further evolve this process to consider image difficulty in relation to task difficulty (e.g. fine detail differentiation is harder than coarse detail differentiation, which can for instance be trivially approximated with hierarchical datasets). Shrivastava et al. \cite{hem}, on the other hand, in their hard example mining paper, rank the images in order of difficulty at the time of training to dynamically generate a mini-curriculum at each iteration.
	
	Kumar et al. \cite{emssvm}, in their work on self-paced learning, instead monitor image difficulty as either the negative log-likelihood for expectation-maximisation or the upper bound on risk for latent structural support vector machines. Jiang et al. \cite{spcl} incorporate both self-paced learning and curriculum learning into a single framework. That is, the curriculum is pre-defined by some expert, but takes into account the feedback from the model (the learner) when selecting which images to propose to the network during training.
	
	Finally, Matiisen et al. \cite{curriculumMastery} introduce the concept of {\it mastery} into the curriculum learning framework. In its simplest form, {\it mastery} is reaching a performance threshold for the model, identified by prior expert knowledge. The model is presented with images from a global dataset, but with a higher probability of sampling images from the current curriculum subset. As the model masters this subset, the probability of sampling these images decreases, while the probability of sampling the next curriculum subset increases.
	
	If we consider these studies concurrently, it is evident that curriculum learning and hard example mining both greatly benefit the deep learning optimisation process, and the combination of the two does so even more. We therefore uniquely propose to employ such methods within NAS, specifically, levying {\it mastery} from \cite{curriculumMastery} in tandem with our own hard example mining approach reminiscent of the {\it`instructor---student collaborative'} learning paradigm \cite{spcl}. 
	
	The work of Cazenavette et al. \cite{distillation} builds upon well-explored dataset distillation techniques \cite{distillationoriginal}. By optimising the $l2$ loss of the \textit{parameters} of a network trained on only 50 images per class, compared to optimal network parameters (i.e. parameters induced by training with 5000 images per class), they are able to achieve reasonable performance (71.5\% on CIFAR-10 \cite{cifar10}). On this basis, we can deduce that training on a fraction of images yields a promising research direction, to which our method pertains without such loss in performance.
	
	\section{Proposed Approach} \label{section:definitions}
	\noindent
	In this section, we detail the process by which our proposed DDS-NAS training strategy dynamically samples the dataset in an online fashion within the NAS cycle (Figure \ref{fig:flow}). DDS-NAS is subsequently deployed across three leading contemporary NAS frameworks (DARTS \cite{darts}, P-DARTS \cite{pdarts}, and TAS \cite{tas}). 
	
	Firstly, we define some key terms to which we will refer in our subsequent discussion:
	\begin{itemize}
		\setlength\itemsep{-0.1cm}
		\item  {\it hard} or {\it hard}-ness: a given example within the dataset at the current NAS training cycle iteration is defined as being {\it hard} if the output of the current model correlates poorly with the ground truth label for this example and hence contributes significantly to the current loss value for the model (i.e. it is either misclassified or classified with a low confidence score in the context of image classification).
		\item  {\it easy}: the converse of {\it hard}, where for a given example the output of the current model correlates strongly with the ground truth label for this example and hence contributes less significantly to the current loss value for the model (i.e. correctly classified with a high confidence score in the context of image classification).
		\item  {\it mastery}: a measure of when a given {\it a priori} performance threshold is reached on the current data subset such that the number of {\it easy} examples in the dataset is high with regard to the current model. 
	\end{itemize}
	
	\subsection{Curriculum Learning Within NAS} \label{subsection:method:cl}
	\noindent
	To formulate an unbiased subset of the global dataset, we use the hard example mining process detailed in Section \ref{subsection:method:hem}. At every training iteration within the NAS search phase, we present such a subset to the NAS model. Following the success of \cite{curriculumMastery}, we in fact present the same subset until it has been \textit{mastered}, according to some \textit{a priori} mastery threshold (see Section \ref{subsection:exp:nas}). Only when the NAS model masters a subset do we sample a new set of examples from the global dataset. If the \textit{mastery} threshold is very low, this subset of data will change often. If the mastery threshold is very high, a given subset is presented to the NAS model for several successive iterations, and a smaller portion of the global dataset is sampled throughout the entire training process. Akin to the restriction with P-DARTS \cite{pdarts} whereby only network parameters (i.e. weights) are updated and  not architectural parameters within the first 10 training epochs, we similarly restrict DDS-NAS from resampling the dataset in this way for the first 10 epochs of NAS training.

	\subsection{Dynamic Data Selection} \label{subsection:method:hem}
	\noindent
	In order to both minimise the data subset used in each NAS iteration without performance degradation and facilitate efficient inter-iteration dataset resampling, we require a low-overhead process by which we can dynamically select new data examples. 
	
	From the initial NAS training iteration, and the immediate subsequent iterations thereafter, model performance can be considered near-random.\footnote{Noting that Deep Image Priors \cite{dip} indicate that untrained model performance in fact correlates to architecture design.} As such, we necessarily depend upon a resampling process independent of model performance, and hence propose the use of dataset example similarity as an alternative measure to relative {\it hard}-ness between samples. The intuition is that a model will perform poorly on examples with greater dissimilarity to those upon which it has already been trained. By using a resampling process independent of model performance, we do not need to compute the forward-pass of the model on all image samples in the entire dataset per hard example mining iteration, an approach commonplace among existing hard example mining approaches. This significantly reduces the computational complexity of DDS-NAS.
	
	Given the need to perform efficient {\it one-to-many} feature distance comparisons via an online approach, we construct a series of efficient furthest-neighbour \textit{kd}-tree structures from the chosen $N$-dimensional feature representations of each example in our global dataset. In order to maintain a balanced data subset in the presence of dynamic reselection, we construct one such \textit{kd}-tree structure per class label in the dataset, resulting in $m$ trees for $m$ dataset classes. In this way we can facilitate {\it like-for-like} class-aware resampling and hence maintain dataset balance throughout the NAS training cycle. This strategy resembles undersampling, which has been shown to be effective for dealing with biased datasets \cite{surveyimbalance}, and is a significant advantage of our approach.
	
	
	To enable efficient look-up within our \textit{kd}-tree structure, we require a sufficiently low dimension $N$ of our feature representation such that the approximate furthest neighbour algorithm does not collapse \cite{nnbreak}. As the dimensionality of image data is high (i.e. $N=28\times28$ in case of MNIST \cite{mnist}, and larger for more complex datasets), we instead propose using an additional autoencoder architecture to construct an image similarity embedding with a much lower dimension ($N=8$ for easier MNIST and Fashion-MNIST datasets \cite{fashion}, $N=32$ for CIFAR-10).
	
	In general, we find that contemporary state-of-the-art autoencoder architectures \cite{skipganomaly} employ skip-connections between the encoder and decoder sub-networks to facilitate improved image reconstruction. However, in this instance, such skip connections are detrimental to the performance of the encoder network in terms of constructing an encoding at the bottleneck of the encoder-decoder architecture (our embedding) that maximally captures the highest level of feature detail within itself. On this basis,  we employ the proven autoencoder architecture from GANomaly \cite{ganomaly} as it is one of the most successful encoder-decoder architectures employed for encoded image discrimination, predating the wider move to the use of skip-connections in the field \cite{skipganomaly}. 
	
	We require that the use of this encoder architecture results in a compact feature embedding that retains the property of spatial similarity such that similar images have similar embeddings within the latent space and vice versa. This property must not come at the expense of image reconstructability. Otherwise, we cannot be confident that a given embedding represents a given image. In other words, there would be no correlation between embedding space dissimilarity and image space dissimilarity. Given reconstructability without similar images clustering within the embedding space, we cannot guarantee that the correlation is strong. 
	
	To enforce these properties, we discovered that contractive loss \cite{contractive} is sufficient for easier datasets, while harder datasets require a combined triplet margin ranking loss with MSE reconstruction loss, weighted via Kendall Loss \cite{kendall}. Subsequently, we can thus order images by their dissimilarity within our furthest neighbour \textit{kd}-tree structures. See Table \ref{tab:autoencoder_hyperparameters} for a lightweight autoencoder training configuration sufficient for each dataset.
	
	During a given NAS training iteration, we measure the {\it hard}-ness of each example image in the current data subset based on cross-entropy loss, following our earlier definition of {\it hard} and {\it easy} examples. To subsequently update our data subset in a dynamic manner,  we first retain the images that are {\it hard} when averaged across the most recent epochs, according to some {\it a priori} {\it hard}-ness threshold (see Section \ref{subsection:exp:nas}). Secondly, by selecting the \textit{kd}-tree from our set that is associated to the class label of each image in the current data subset below the {\it hard}-ness threshold (i.e. the {\it easy} images), we can then identify the most dissimilar image of the same class in the global training set in $O(log(n))$ time. We can then use this to replace the {\it easy} image  within the data subset. This dynamically updated training data subset will then be used for the next NAS training iteration. A detailed example can be found in Appendix \ref{app:sec:hem}.

	Our overall pipeline is presented as follows: once the previous data subset has been {\it mastered} by iterative NAS training, we dynamically formulate a new balanced subset of the global training dataset based on (a) the retention of images that are considered {\it hard}, and (b) the replacement of images that are considered {\it easy} with dissimilar images of the same class to retain dataset balance (Fig. \ref{fig:flow}). Pseudo-code to illustrate the overall pipeline with time complexity analysis can be found in Appendix \ref{app:sec:pseudo}, which highlights the potential search speed efficiency that DDS-NAS affords.

	\begin{figure}[!ht]
		\small
		\begin{tabular}[t]{cc}
			\begin{subfigure}[b]{0.45\linewidth}
				\includegraphics[width=\linewidth, trim={2.75cm 1.75cm 0cm 2cm}, clip]{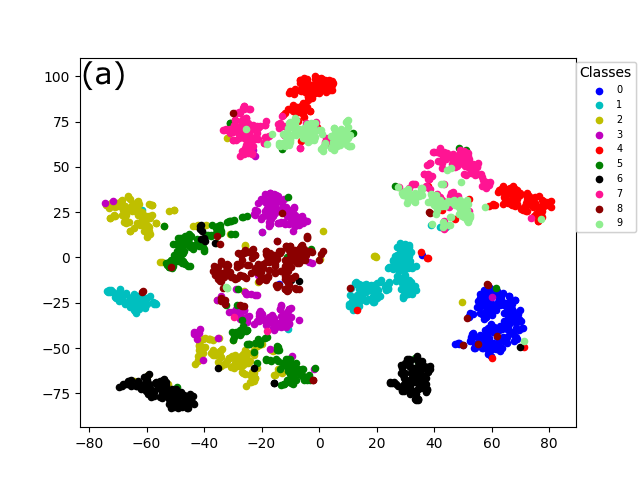}
				\centering
				\label{fig:mnist}
			\end{subfigure}&
			\begin{subfigure}[b]{0.45\linewidth}
				\includegraphics[width=\linewidth, trim={2.75cm 1.75cm 0cm 2cm},clip]{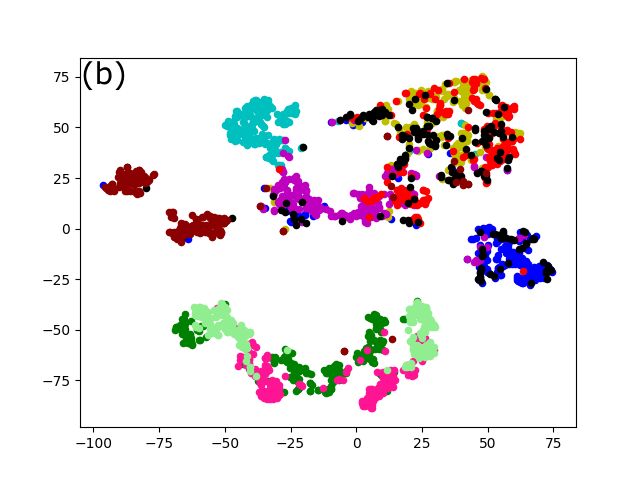}
				\centering
				\label{fig:fashion}
			\end{subfigure}\\
			\vspace{-0.5cm}
			\begin{subfigure}[b]{0.45\linewidth}
				\includegraphics[width=\linewidth, trim={2cm 1.25cm 0cm 1cm},clip]{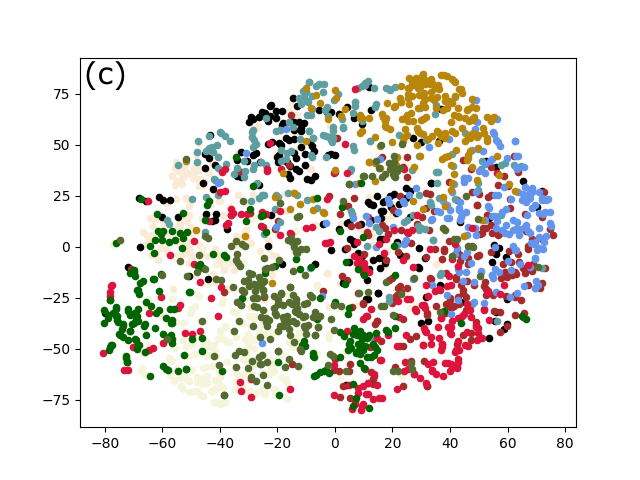}
				\centering
				\label{fig:cifar10}
			\end{subfigure}&
			\begin{subfigure}[b]{0.45\linewidth}
				\includegraphics[width=\linewidth, trim={2.75cm 1.75cm 0cm 0cm}, clip]{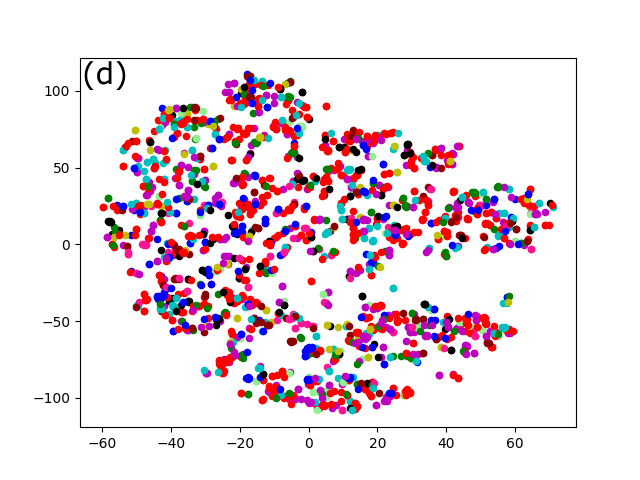}
				\centering
				\label{fig:planes}
			\end{subfigure}
		\end{tabular}
		\caption{TSNE visualisation of clustering of autoencoded image feature representation within latent space. Our autoencoder preserves the property that similar images have similar encodings for MNIST (a), Fashion-MNIST (b), and CIFAR-10 (c). However, our compact embedding  is unsuitable for fine-grained image classification such as FGVC-Aircraft (d), which is a known limitation of autoencoders.}
		\label{fig:clustering}
	\end{figure}

	\begin{table}[h!]
		\centering
		\begin{tabular}{|c|c|c|c|c|}
			\hline
			Dataset & \makecell{Autoencoder\\Architecture} & \makecell{Bottleneck\\Embedding\\Dimension N} & Loss Function\\
			\hline
			MNIST & GANomaly & 8 & Contractive Loss\\
			\hline
			Fashion-MNIST & GANomaly & 8 & Contractive loss\\
			\hline
			CIFAR-10 & GANomaly & 32 & \makecell{Triplet loss and\\MSE loss with\\ Kendall Loss Weighting}\\
			\hline
		\end{tabular}
		\caption{Suggested autoencoder training configuration parameters for each dataset to yield a sufficiently lightweight architecture that can generate low-dimensionality embeddings.}
		\label{tab:autoencoder_hyperparameters}
	\end{table}
		
	\section{Experimental Setup}
	\noindent
	We detail our experimental setup for DDS-NAS deployment across the Differentiable Architecture Search (DARTS), Progressive DARTS (P-DARTS) and Network Pruning via Transformable Architecture Search (TAS) NAS frameworks. This setup is used to demonstrate the performance of our proposed approach with several image classification datasets. 
	
	\subsection{NAS Configuration} \label{subsection:exp:nas}
	\noindent	
	Unless otherwise stated, all employed NAS frameworks adopt the same common configuration using Adam optimisation \cite{adam} with initial learning rate $lr = 3e^{-4}$, weight decay $wd = 1e^{-3}$, and momentums $\beta_{1}=0.5$ and $\beta_{2}=0.999$ (P-DARTS uses $lr=6e^{-4}, wd=1e^{-3}$, TAS uses $lr = 1e^{-4}$). For weight optimisation for the NAS-derived architectures themselves, we use an SGD optimiser with $wd = 3e^{-4}$, and momentum $\beta=0.9$ (P-DARTS uses $wd=5e^{-4}$). Additionally, for DARTS we employ a Cyclic Learning Rate Scheduler with base $lr = 0.001$, max $lr = 0.01$, and step size up $=$ step size down $= 10$. We set $lr = 0.01$ when the previous dynamically selected data subset is mastered, and an updated data subset is introduced. Therefore, the updated data subset is learned quickly and is then `fine-tuned' as with the previous subset. There is precedence for such an approach in SGDR \cite{sgdr}, in which the learning rate is periodically reset to a higher value before the learning rate decay is reapplied. P-DARTS and TAS both adopt Cosine Annealing Learning Rate Scheduler with $lr=2.5e^{-2}$ and $lr=0.1e^{-2}$ respectively. We select the ResNet-110 architecture for TAS $kd$-teacher training.
	The models are implemented using PyTorch \cite{pytorch} (v1.6.0, Python 3.6.9).
	
	Performance of DDS-NAS deployed across each NAS framework is presented in terms of both Top-1 accuracy and parameter count (complexity) of the optimal NAS-generated architecture, together with the computational effort of the NAS search phase (in GPU days) across all three datasets.
	
	Experimentation indicates that our NAS framework is generally insensitive to \textit{a priori} thresholds that do not need to be exhaustively searched. A subset-size of 100 is sufficient for the easier MNIST \cite{mnist} and Fashion-MNIST \cite{fashion} tasks, and 1000 for CIFAR-10 \cite{cifar10}. Adopting a high hardness threshold (\textit{hard}-ness threshold $> 0.8$) across all datasets and all NAS strategies enables the searched network architecture to formulate a thorough feature representation for image classification. The best network architectures are discovered with a mastery threshold $\approx 0.5$. P-DARTS and TAS learn deep representations for images slower than DARTS. This can be attributed to the additional tasks done alongside reducing classification loss, wherein P-DARTS progressively restricts the search space while increasing architecture depth, and TAS minimises for network architecture complexity. Conversely, DARTS can afford a lower mastery threshold ($\approx 0.15$) for the easier MNIST and Fashion-MNIST tasks, but the performance gain is marginal. All presented results use the same hardness ($0.85$) and mastery ($0.5$) thresholds to ensure fairness.
	
	\subsection{Hard Example Mining}
	\noindent	
	The GANomaly autoencoder \cite{ganomaly} used to encode the images into their latent space representation is trained with Contractive Loss \cite{contractive} for 30 epochs, with $bs=8$, and Adam optimiser with momentums $\beta_{1}=0.9$ and $\beta_{2}=0.999$, $wd=0$, $lr=1e^{-3}$. For the CIFAR-10 task, the autoencoder is instead trained with combined triplet margin loss \cite{triplet} and MSE reconstruction loss, weighted under Kendall Loss \cite{kendall}.
	
	\section{Evaluation}
	\noindent
	Having validated the feature representation embedding that underpins our dynamic data selection via hard example mining (see Figure \ref{fig:clustering}), we present out evaluation in terms of DDS-NAS comparison to contemporary state-of-the-art approaches, with supporting ablation studies.
	
	\begin{table*}[!t]
		\scriptsize
		\begin{center}
			\renewcommand*{\arraystretch}{2}
			\resizebox{\textwidth}{!}{%
			\begin{tabular}{|c|c|c|c|c|c|}
				\hline
				Dataset & NAS Approach & \makecell{Top-1 Accuracy (\%) $\uparrow$ \\ DARTS / \\ \textcolor{blue}{P-DARTS} / \textcolor{green}{TAS}} & \makecell{Params (M) $\downarrow$  \\ DARTS / \\ \textcolor{blue}{P-DARTS} / \textcolor{green}{TAS}} & \makecell{Search Cost (GPU Days) $\downarrow$\\ DARTS / \\ \textcolor{blue}{P-DARTS} / \textcolor{green}{TAS}}\\
				\hline
				\multirow{2}{*}{\rotatebox[origin=c]{90}{MNIST}} & Original & 99.75 / \textcolor{blue}{99.26} / \textcolor{green}{99.27$\dagger$} & \underline{0.66} / \textcolor{blue}{3.68} / \textcolor{green}{1.00} & 0.51 / \textcolor{blue}{1.89} / \textcolor{green}{0.28}\\\cdashline{2-6}
				& DDS-NAS & \underline{99.78} / \textcolor{blue}{99.17} / \textcolor{green}{99.30$\dagger$} & 0.75 / \textcolor{blue}{3.51} / \textcolor{green}{0.81} &  0.030 / \textcolor{blue}{0.070} / \textcolor{green}{\underline{0.021}}\\\hline
				\multirow{2}{*}{\rotatebox[origin=c]{90}{\makecell{Fashion\\MNIST}}} & Original & 95.33
				/ \textcolor{blue}{93.42} / \textcolor{green}{95.09$\dagger$} & 3.27 / \textcolor{blue}{4.04} / \textcolor{green}{0.94} & 0.63 / \textcolor{blue}{1.98} / \textcolor{green}{0.27}\\\cdashline{2-6}
				& DDS-NAS  & \underline{95.48} / \textcolor{blue}{93.04} / \textcolor{green}{95.08$\dagger$} & 3.44 / \textcolor{blue}{4.23} / \textcolor{green}{\underline{0.83}}& \underline{0.030} / \textcolor{blue}{0.078} / \textcolor{green}{0.031}\\
				\hline
				\multirow{6}{*}{\rotatebox[origin=c]{90}{CIFAR-10}} & Original & \underline{97.17} / \textcolor{blue}{96.50}
				/ \textcolor{green}{93.89} & 3.16 / \textcolor{blue}{3.43} / \textcolor{green}{\underline{0.85}} & 1.78 / \textcolor{blue}{0.65} / \textcolor{green}{0.26}\\\cdashline{2-6}
				& DDS-NAS  & 96.57 / \textcolor{blue}{95.07} / \textcolor{green}{93.12} & 3.72 / \textcolor{blue}{4.13} / \textcolor{green}{1.06} & 0.36 / \textcolor{blue}{0.095} / \textcolor{green}{\underline{0.040}}\\\cline{2-6}
				& Shapley-NAS \cite{shapley} & 96.96 & 3.60 & 0.36 \\\cdashline{2-6}
				& SNAS \cite{snas} & 97.15 & 2.85 & 1.83 \\\cdashline{2-6}
				& DenseNet \cite{densenet} & 94.23 & 7.0 & -- \\
				\hline
			\end{tabular}}
		\end{center}
		\caption{Accuracy, memory footprint, and (search-phase) training cost of final generated model from DDS-NAS deployed upon DARTS, P-DARTS, and TAS, compared to their original implementations and others. $\dagger$ indicates results without $kd$-teacher training owing to the lack of available teacher models for MNIST and Fashion-MNIST datasets.}
		\label{tab:mastery}
	\end{table*}

	\begin{table*}[!t]
		\scriptsize
		\begin{center}
			\renewcommand*{\arraystretch}{2}
			\resizebox{\textwidth}{!}{%
			\begin{tabular}{|c|c|c|}
				\hline
				\multicolumn{3}{|c|}{CIFAR-100} \\\hline
				
				NAS Approach & \makecell{Top-1 Accuracy  (\%) $\uparrow$ \\ DARTS / \textcolor{blue}{P-DARTS} / \textcolor{green}{TAS}} & \makecell{Params (M) $\downarrow$ \\ DARTS / \textcolor{blue}{P-DARTS} / \textcolor{green}{TAS} } \\\hline
				Original & \makecell{81.33 / \textcolor{blue}{80.58}
				/ \textcolor{green}{71.72}} & \makecell{2.75
				/ \textcolor{blue}{3.49}
				/ \textcolor{green}{1.15}} \\\cdashline{1-3}
				DDS-NAS & \makecell{82.64 / \textcolor{blue}{75.45} / \textcolor{green}{70.47}} & \makecell{3.80 / \textcolor{blue}{4.24}
				/ \textcolor{green}{1.15}} \\\cline{1-3}
				Shapley-NAS & 83.42 & 3.66 \\\cdashline{1-3}
				DenseNet & 76.21 & 7.0 \\\hline
				
				\multicolumn{3}{|c|}{ImageNet}\\\hline
				Original & 73.30 / \textcolor{blue}{75.72}
				/ \textcolor{green}{-} & 4.51
				/ \textcolor{blue}{4.94}
				/ \textcolor{green}{-} \\\cdashline{1-3}
				DDS-NAS & 76.26 / \textcolor{blue}{75.63} / \textcolor{green}{-} & 6.14 / \textcolor{blue}{5.68}
			/ \textcolor{green}{-}\\\cline{1-3}
				Shapley-NAS & 75.52 & 5.14  \\\cdashline{1-3}
				DenseNet & 74.98 & 7.0 \\\hline
			\end{tabular}}
		\end{center}
		\caption{Accuracy and memory footprint of CIFAR-10 searched models transferred to CIFAR-100 and ImageNet}
		\label{tab:cifar100}
	\end{table*}

	\subsection{Neural Architecture Search}
	\label{subsection:eval:nas}
	\noindent
	Table \ref{tab:mastery} presents the performance obtained by the final model generated by DDS-NAS with respect to each dataset under consideration. Across all cases, the performance of our generated models is competitive with the state of the art, with minimal to no impact on generated model size. Moreover, across all cases, we substantially lower the computational efforts required for NAS (0.07 GPU days compared to 1.89 in the case of P-DARTS for MNIST, \underline{ 27 times quicker})\footnote{Still an order of magnitude faster even after factoring in the time taken to train the autoencoder}. Since we can determine a replacement image for our dynamic subset in average case $O(log(n))$ time, we are able to \ul{reduce the search phase training cost by one order of magnitude} over state-of-the-art results.
	
	Without loss in performance, our hard example mining method yields discriminative architectures that can be transferred to CIFAR-100 \cite{cifar10} and ImageNet \cite{imagenet} (Table \ref{tab:cifar100}). However, reproducibility presents a particularly significant problem within the NAS domain \cite{nasreproducability} and TAS ImageNet performance is considerably lower than the literature reports. DDS-NAS-TAS performance is omitted for fairness. Whilst our technique is demonstrated upon commonplace NAS approaches (DARTS, P-DARTS, TAS) it could equally be deployed on top of more recent advancements \cite{shapley, dnal, senas}, further minimizing any difference in performance.
	
	\subsection{Ablation Studies}
	\label{subsection:eval:ablation}
	\begin{table*}[!t]
		\scriptsize
		\begin{center}
			\renewcommand*{\arraystretch}{2}
			\resizebox{\textwidth}{!}{%
			\begin{tabular}{|c|c|c|c|c|c|}
				\hline
				Dataset & NAS Approach & \makecell{Top-1 Accuracy\\(Search Phase) (\%) $\uparrow$ \\ DARTS / \\ \textcolor{blue}{P-DARTS} / \textcolor{green}{TAS}} & \makecell{Top-1 Accuracy\\ (Final) (\%) $\uparrow$ \\ DARTS / \\ \textcolor{blue}{P-DARTS} / \textcolor{green}{TAS}}& \makecell{ Params (M) $\downarrow$ \\ DARTS / \\ \textcolor{blue}{P-DARTS} / \textcolor{green}{TAS}}\\
				\hline
				\multirow{3}{*}{\rotatebox[origin=c]{90}{MNIST}} & DDS-NAS & 94.00 / \textcolor{blue}{78.89} / \textcolor{green}{44.89} & 99.78 / \textcolor{blue}{99.17} / \textcolor{green}{99.30} & 0.75 / \textcolor{blue}{3.51} / \textcolor{green}{0.81}\\\cdashline{2-6}
				& \makecell{Original framework with \\dataset size 100 }& 78.28 / \textcolor{blue}{70.81} / \textcolor{green}{39.24} & 94.43 / \textcolor{blue}{98.69} / \textcolor{green}{99.18} &  0.70 / \textcolor{blue}{4.54} / \textcolor{green}{0.53}\\\cdashline{2-6}
				& \makecell{DDS-NAS with \\untrained autoencoder} & 92.28 / \textcolor{blue}{78.76} / \textcolor{green}{51.96} & 95.28 / \textcolor{blue}{98.78} / \textcolor{green}{99.21} & 0.75 / \textcolor{blue}{3.11} / \textcolor{green}{1.05}\\
				\hline
				\multirow{3}{*}{\rotatebox[origin=c]{90}{Fashion-MNIST}} & DDS-NAS & 72.92 / \textcolor{blue}{65.71} / \textcolor{green}{32.71} & 95.48 / \textcolor{blue}{93.04} / \textcolor{green}{95.08} & 3.44 / \textcolor{blue}{4.23} / \textcolor{green}{0.83}\\\cdashline{2-6}
				& \makecell{Original framework with \\dataset size 100 }  & 56.27 / \textcolor{blue}{58.16} / \textcolor{green}{35.66} & 91.69 / \textcolor{blue}{90.03} / \textcolor{green}{94.61} & 3.47 / \textcolor{blue}{4.69} / \textcolor{green}{0.46}\\\cdashline{2-6}
				& \makecell{DDS-NAS with \\untrained autoencoder} & 69.49 / \textcolor{blue}{64.47} / \textcolor{green}{39.12}& 92.04 / \textcolor{blue}{91.52} / \textcolor{green}{94.87}&  3.48 / \textcolor{blue}{3.92} / \textcolor{green}{0.93}\\
				\hline
				\multirow{3}{*}{\rotatebox[origin=c]{90}{CIFAR-10}} & DDS-NAS  & 56.00 / \textcolor{blue}{22.14} / \textcolor{green}{29.88} & 96.57 / \textcolor{blue}{95.07} / \textcolor{green}{93.12}& 3.72 / \textcolor{blue}{4.13} / \textcolor{green}{1.06}\\\cdashline{2-6}
				& \makecell{Original framework with \\dataset size 1000 }  & 51.02 / \textcolor{blue}{41.70} / \textcolor{green}{23.81} & 88.58 / \textcolor{blue}{85.74} / \textcolor{green}{90.83} & 3.55 / \textcolor{blue}{4.04} / \textcolor{green}{0.32}\\\cdashline{2-6}
				& \makecell{DDS-NAS with \\untrained autoencoder} & 51.10 / \textcolor{blue}{46.59} / \textcolor{green}{28.21}& 88.90 / \textcolor{blue}{88.96} / \textcolor{green}{91.72} & 3.64 / \textcolor{blue}{4.25} / \textcolor{green}{0.83}\\
				\hline
			\end{tabular}}
		\end{center}
		\caption{Ablation studies: accuracy and memory footprint of models generated by DDS-NAS, models generated by the original framework with limited data (equivalent to removing hard example mining and curriculum learning), and models generated by DDS-NAS with an untrained autoencoder (equivalent to removing hard example mining).}
		\label{tab:ablation}
	\end{table*}
	
	\noindent
	To validate our proposed approach, we compare the performance of DDS-NAS to selected NAS frameworks, both: (a) without \textit{dynamic data selection} in order to ablate the contribution of our combined hard example mining and curriculum learning  strategy; and (b) with an untrained autoencoder to ablate the contribution of the image-dissimilarity based hard example mining strategy.
	
	\subsubsection{Without Dynamic Data Selection} \label{subsubsection:eval:pipeline}
	\noindent
	For each dataset, we employ all three original implementations (DARTS \cite{darts}, P-DARTS \cite{pdarts}, TAS \cite{tas}), but with a subset of the data at each training iteration. \textit{This is equivalent to omitting both hard example mining and curriculum learning}. We use the same volume of data as adopted by DDS-NAS: 100 randomly selected images for MNIST and Fashion-MNIST, and 1000 for CIFAR-10. Subsequently, we can determine the impact of our curriculum learning and hard example mining pipeline. Comparing the first and second row of the results for each dataset presented in Table \ref{tab:ablation}, it is evident that DDS-NAS achieves substantially improved accuracy while yielding fractionally larger architectures in some cases. This behaviour is exhibited in MNIST, where the original DARTS framework achieves only 78.28\% accuracy after the search phase, and 94.43\% accuracy after fine-tuning the stacked searched cell (compared to 94.00\% and 99.78\% respectively for DDS-NAS-DARTS). This performance difference is further highlighted with both the other datasets and other frameworks. The final performance of the original P-DARTS implementation falls behind DDS-NAS across all datasets (85.74\% compared to 95.07\% for CIFAR-10, for instance). Interestingly, with hard example mining and curriculum learning omitted in this manner, TAS generates smaller models (0.32M compared to 1.06M for CIFAR-10), but often at the expense of accuracy.
	
	\subsubsection{Untrained Autoencoder} \label{subsubsection:eval:autoencoder}
	\begin{figure}[!ht]
		\small
		\begin{tabular}[t]{cc}
			\begin{subfigure}[b]{0.45\columnwidth}
				\includegraphics[width=\linewidth, trim={2cm 1.25cm 1.25cm 1.25cm}, clip]{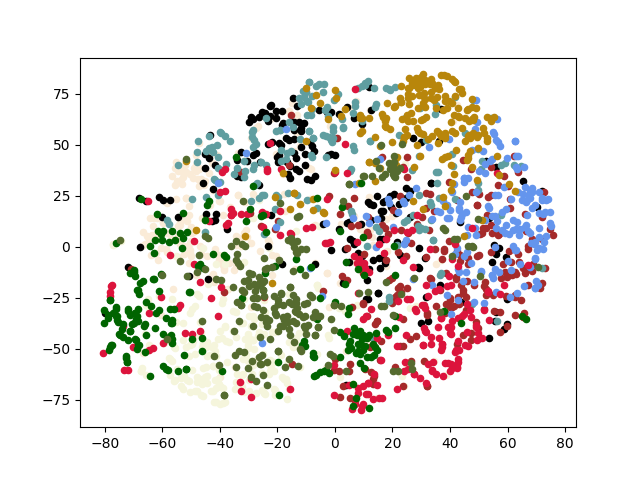}
				\centering
				\label{fig:cifar10triplet}
			\end{subfigure}
			&
			\begin{subfigure}[b]{0.45\columnwidth}
				\includegraphics[width=\linewidth, trim={2cm 1.25cm 1.25cm 1.25cm}, clip]{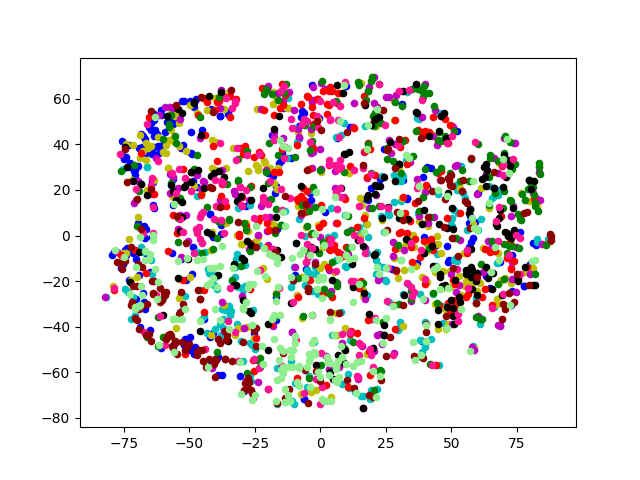}
				\centering
				\label{fig:cifar10cae}
			\end{subfigure}
		\end{tabular}
		\caption{TSNE visualisation of the clustering of autoencoded CIFAR-10 image feature representation within the latent space. Training with triplet margin loss with Kendall loss achieves good clustering (left). Training with contractive loss achieves poor clustering (right).}
	\label{fig:tripletclustering}
	\end{figure}

	We ablate the autoencoder-derived feature embedding within our hard example mining method by replacing the DDS-NAS autoencoder with one that is untrained, and thus unable to determine the most dissimilar images from a given training data subset. \textit{This can be considered as a process equivalent to curriculum learning without hard example mining}, as the images are effectively randomly sampled. This time, we compare the first and third row for each dataset in Table \ref{tab:ablation}. Evidently, the models generated by DDS-NAS with an untrained autoencoder are significantly worse (for instance 92.04\% compared to 95.48\% upon Fashion-MNIST by DDS-NAS-DARTS). On this basis, we can therefore conclude that DDS-NAS necessarily requires a suitable hard example mining approach, for which our image similarity strategy is sufficient. 
	
	Furthermore, an autoencoder that achieves good reconstruction but a mediocre clustering of embedded features is inadequate for DDS-NAS (Fig. \ref{fig:tripletclustering}, Table \ref{tab:ablauto}). Bad clustering and thus ineffective hard example mining yields inferior classification accuracy (95.29\%) compared to hard example mining with good clustering (96.57\%). Similarly, sufficient clustering but poor reconstruction is detrimental to DDS-NAS (94.94\%). Lack of both properties yields significantly worse performance (88.90\%), wherein there is no correlation between embedding space dissimilarity and image space dissimilarity at all.

	\begin{table}[!t]
		\scriptsize
		\begin{center}
			\renewcommand*{\arraystretch}{2}
			\begin{tabular}{|c|c|c|c|c|}
				\hline
				Reconstruction & \checkmark & \checkmark & $\times$ & $\times$ \\\hline
				Clustering & \checkmark & $\times$ & \checkmark & $\times$ \\\hline
				Top-1 Accuracy (\%) $\uparrow$ & 96.57 & 95.29 & 94.94 & 88.90 \\\hline
			\end{tabular}
		\end{center}
		\caption{Accuracy of DDS-NAS-DARTS employing autoencoders with different capabilities on CIFAR-10.}
		\label{tab:ablauto}
	\end{table}
	
	By comparing row two (neither hard example mining nor curriculum learning) and row three (curriculum learning but not hard example mining) for each dataset in Table \ref{tab:ablation}, it is clear that our curriculum learning methodology is \textit{somewhat} effective even without incorporating hard example mining. DDS-NAS performance with an untrained autoencoder exceeds that of the original framework with limited data in all cases (88.58\% compared to 88.90\% for CIFAR-10 with DARTS, 90.03\% compared to 91.52\% for Fashion-MNIST with P-DARTS).
	
	\section{Limitations}
	The modularity of the proposed DDS-NAS framework provides a significant advantage over existing NAS methods, and allows it to be adopted alongside multiple NAS frameworks. Selecting an off-the-shelf autoencoder or training one from scratch is a reasonable approach provided it can generate a low-dimensionality embedding space that offers reasonable reconstruction and clustering capabilities (see Section \ref{subsection:eval:ablation}). For fine-grained classification tasks however, this is a challenge (see Figure \ref{fig:clustering}) and remains an open area of research. 
	
	In addition, the current DDS-NAS approach requires one \textit{kd}-tree per class so that we can perform class-aware dynamic dataset updates. While this offers reasonable robustness towards biased datasets, long-tailed distributions in datasets may present additional challenges, where there are not enough samples for a given class. We might expect training samples to be memorized in this situation, yielding noisy architecture weight-update steps. One simple solution to resolve this might be to combine samples from classes with few samples into a single \textit{kd}-tree but this is a direction for future research.
	
	\section{Conclusion}
	To conclude, we propose DDS-NAS: a novel NAS framework capable of reducing the time required for the NAS search phase by one order of magnitude. By employing image similarity as a basis for hard example mining, and thus (online) dynamic data sub-selection, DDS-NAS yields models that remain competitive towards accuracy and memory costs upon common image datasets. Further, we demonstrate that DDS-NAS can be deployable upon several NAS approaches and architecture spaces, and is similarly extendable to all existing evolutionary, reinforcement-learning, or gradient-based NAS approaches. DDS-NAS can even incorporate NAS search phase techniques that are deployed alongside rather than in place of existing NAS approaches \cite{dnal}.
	
	Following the success of our approach, we posit that only a fraction of commonly used image datasets contribute to learning. As such, additional analysis of these datasets is necessary. Moreover, a more comprehensive investigation into the autoencoder architecture employed within our hard example mining method may yield better results (and thus further reduce the volume of contributing images within a dataset). Specifically, we require an autoencoder that can generate similar embeddings for similar images even within the fine-grain classification domain. Alternative autoencoder losses, or measures of image similarity such as hashing, may yield similar improvements. 
	Nevertheless, even a glimpse of image similarity as a metric within hard example mining has proven extremely effective. We thus introduce several new avenues for improvement, particularly alongside NAS frameworks, and demonstrate that network architecture topology design should not necessarily be the sole consideration for future NAS solutions.
	
	\section{Declaration of generative AI and AI-assisted technologies in the writing process}
	During the preparation of this work the authors used GPT-4o mini in order to generate the LaTeX code for Algorithm \ref{algorithm:overall} with placeholder steps and time complexity. After using this tool/service, the authors reviewed and edited the content as needed and take full responsibility for the content of the published article.
	
	\section{Acknowledgements}
	This work was supported by Durham University, United Kingdom, the European Regional Development Fund Intensive Industrial Innovation Grant No. 25R17P01847 and Petards Joyce-Loebl Ltd.
	\clearpage
	{\small
		\bibliographystyle{elsarticle-num}
		\bibliography{nasreview.bib}
	}

	\begin{appendices}
		\section{Hard Example Mining Search Process} \label{app:sec:hem}
			\begin{figure}[!ht]
				\centering

				\includegraphics[trim={6cm 0 6cm 0},width=\textwidth]{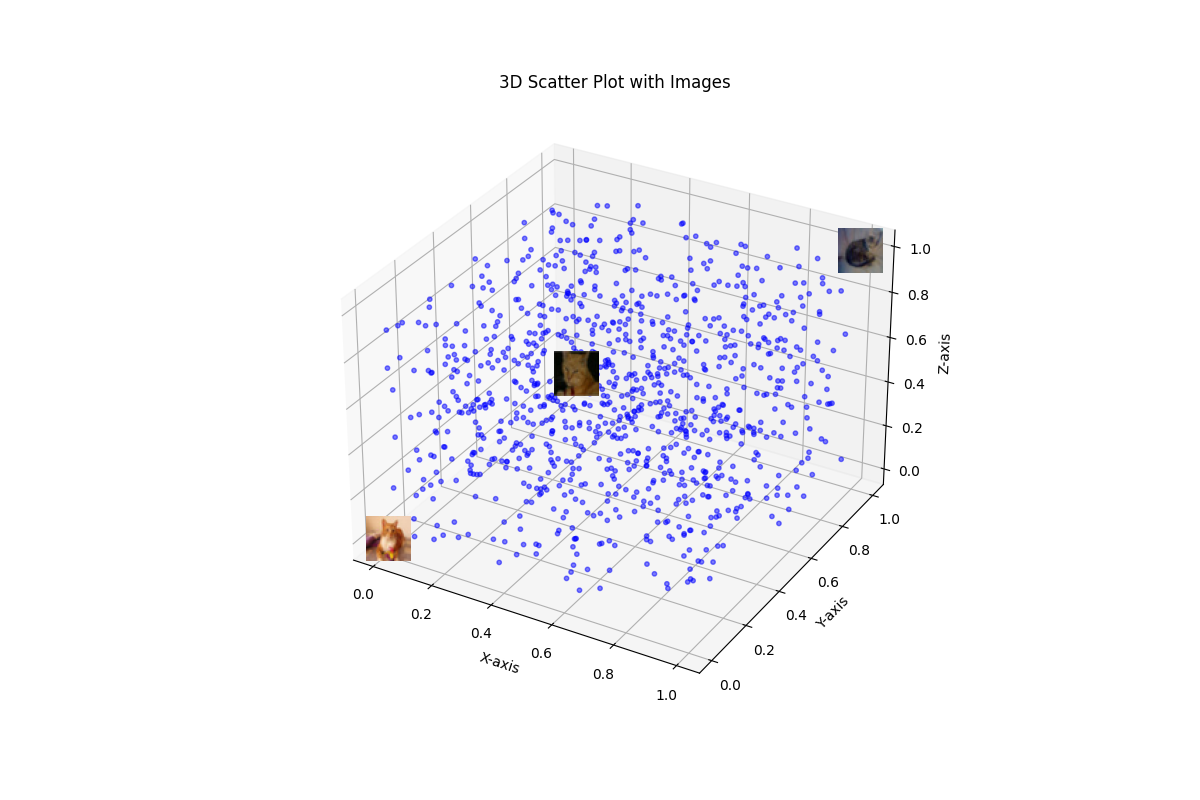}
				\caption{An example embedding space for the cat class.}
				\label{fig:appendix1cat}
			\end{figure}
		
		In the first iterations of training, all samples are considered equally difficult and images from each class are randomly sampled. The embedding space for a cat class might resemble Figure \ref{fig:appendix1cat}. The ginger cat in daylight (coordinates $=0,0,0$) is sampled first, and after a few iterations, the current subset of data is mastered. Performance on the ginger cat sample is good and it is considered \textit{easy}. The most dissimilar image in a given dimension to the current cat sample is the ginger cat in darkness (coordinates $=0,1,0$). The model is trained for a few more epochs and eventually the data subset is mastered again, and performance on the cat in darkness is good. This time the most dissimilar image in a new dimension is the black cat (coordinates $=1,1,1$), which is now added to the data subset and \textit{replaces} the ginger cat. In this way, different features of the cat embedding space are explored during training. In practice, dimensions are often more abstract based on the learnt embedding space of the autoencoder.

		Traditional hard example mining strategies compute hardness for all samples in the entire dataset and re-weight their impact accordingly. By using image dissimilarity as a measure for image hardness, we do not need to evaluate current model performance on images outside of a data subset. We compute similarity embeddings \textit{once} prior to any NAS training iterations and use the embeddings to calculate maximally dissimilar image samples from the current data subset for the next training iteration. We are therefore able to levy the benefits of curriculum learning and hard example mining while \textit{reducing} overhead. This has the benefit of hard example mining and curriculum learning over existing core-set selection NAS approaches \cite{coresetnas, adaptivenas}, which use neither. In this way, DDS-NAS also benefits from data subset selection compared to curriculum strategies \cite{cnas,close}. 
		
		\clearpage
		\section{Overall Pipeline Pseudo-code with Time Complexity Analysis} \label{app:sec:pseudo}
		
		\noindent\textbf{Note:} In this algorithm, $n'$ represents a subset of size much smaller than $n$, i.e., $n' \ll n$.
		
		\begin{algorithm}
			\caption{Pseudo-code illustrating the overall DDS-NAS pipeline}
			\label{algorithm:overall}
			\begin{algorithmic}[1]
				\Statex \textbf{Step} \hfill \textbf{Big-O Time Complexity}
				\State Initialize data subset: \hfill $\mathcal{O}(n')$
				\Statex \hspace{1em}Randomly select images from each class
				\Statex \hspace{1em}to form a balanced subset of size $n'$
				\State Construct $m$ class-specific \textit{kd}-trees \hfill $\mathcal{O}(m n \log n)$
				\State Compute feature embeddings with autoencoder
				\For{each NAS iteration} \hfill $\mathcal{O}(k)$
				\If{current subset is mastered} \hfill $\mathcal{O}(1)$
				\State Retain hard examples based on threshold \hfill $\mathcal{O}(n')$
				\State Identify dissimilar images via \textit{kd}-tree look-up \hfill $\mathcal{O}(\log n')$
				\State Replace easy images with dissimilar ones \hfill $\mathcal{O}(1)$
				\State Update dataset for next NAS iteration \hfill $\mathcal{O}(n')$
				\EndIf
				\State Perform a normal NAS training iteration
				\If{network has converged} \hfill $\mathcal{O}(1)$
				\State \textbf{Stop} \hfill $\mathcal{O}(1)$
				\EndIf
				\EndFor

			\end{algorithmic}
		\end{algorithm}
		
		\noindent\textbf{Note:} As long as $k$ is sufficiently large, this algorithm significantly reduces convergence time. In modern deep learning, $k$ is typically large enough to ensure this.
	\end{appendices}
\end{document}